\documentclass[10pt, letter, onecolumn]{arxiv}

\usepackage{makecell}
\usepackage{kantlipsum, lipsum}
\usepackage{dm-colors}
\usepackage{amsmath}
\usepackage{pstricks, pst-node}
\usepackage{verbatim}
\usepackage{multirow}
\usepackage{scalerel}
\usepackage{booktabs}
\usepackage{enumitem}
\usepackage{xspace}
\usepackage{bm}
\usepackage{bbm}
\usepackage{mathtools}
\usepackage{soul}
\usepackage{epsfig}
\usepackage{graphicx}
\usepackage{csquotes}
\usepackage{setspace}
\usepackage{colortbl}
\usepackage{threeparttable}
\usepackage{tabularx,ragged2e}
\usepackage{placeins}
\usepackage[hang,flushmargin,symbol]{footmisc}

\usepackage{varioref}
\usepackage[pagebackref=false,breaklinks=false,%
            colorlinks=true,bookmarks=true,citecolor=ourdarkblue,%
            urlcolor=ourdarkblue,linkcolor=ourdarkblue]{hyperref}
\usepackage[noabbrev,capitalize]{cleveref}
\usepackage{etoc}
\usepackage{lineno}

\graphicspath{{figures/}}

\usepackage[misc]{ifsym}

\title{\Large{ReXrank: A Public Leaderboard for AI-Powered Radiology Report Generation}}

\author[1 \Letter]{Xiaoman Zhang}
\author[1]{Hong-Yu Zhou}
\author[1]{Xiaoli Yang}
\author[1]{Oishi Banerjee} 
\author[1]{\\ \vspace{0.1cm} Juli\'an N. Acosta}
\author[1]{Mohammed Baharoon} 
\author[2]{Josh Miller}
\author[2,3]{Ouwen Huang}
\author[1]{Pranav Rajpurkar}

\affil[1]{\normalsize Department of Biomedical Informatics, Harvard Medical School, Boston, MA, USA  \authorcr}
\affil[2]{\normalsize Gradient Health, Durham, NC, USA  \authorcr}
\affil[3]{\normalsize Department of Statistical Science, Duke University, Durham, NC, USA  \authorcr}

\renewcommand{\correspondingauthor}[1]{Corresponding Author: Xiaoman Zhang xiaoman\_zhang@hms.harvard.edu.
}

\begin{document}

\begin{abstract}

AI-driven models have demonstrated significant potential in automating radiology report generation for chest X-rays. However, there is no standardized benchmark for objectively evaluating their performance. 
To address this, we present \textbf{ReXrank} (\textbf{\href{https://rexrank.ai}{https://rexrank.ai}}), a public leaderboard and challenge for assessing AI-powered radiology report generation.
Our framework incorporates \textbf{ReXGradient}, the largest test dataset consisting of \textbf{10,000} studies, and three public datasets (MIMIC-CXR, IU-Xray, CheXpert Plus) for report generation assessment. 
ReXrank employs 8 evaluation metrics and separately assesses models capable of generating only findings sections and those providing both findings and impressions sections.
By providing this standardized evaluation framework, ReXrank enables meaningful comparisons of model performance and offers crucial insights into their robustness across diverse clinical settings.
Beyond its current focus on chest X-rays, ReXrank's framework sets the stage for comprehensive evaluation of automated reporting across the full spectrum of medical imaging.

\end{abstract}

\maketitle

\section{Introduction}

Writing accurate radiology reports from medical images is a critical but complex task, requiring both deep expertise in medical imaging and the ability to accurately interpret and articulate intricate findings. 
The demand for such reports has surged with the rapid advancements in imaging technologies, leading to increased workloads for radiologists, risks of information loss, and longer report turnaround times~\cite{bruls2020workload}.

AI-driven solutions have emerged as a potential answer to these challenges, serving as assistive tools to enhance reporting efficiency and ensure access to high-quality, specialty-level interpretations. 
Medical visual-language models have shown promise in automating the generation of radiology reports from chest X-ray images~\cite{chen2024chexagent,pellegrini2023radialog,chen2023cross,lee2023llm,zhou2024generalist}.
However, as the field of AI-assisted medical reporting rapidly evolves, there is a growing need for standardized benchmarks to objectively assess and compare the performance of these models. 
Existing datasets for chest X-ray report generation, such as MIMIC-CXR~\cite{johnson2019mimic}, are valuable but exhibit limitations that hinder their effectiveness for benchmarking. These datasets frequently suffer from inconsistent data splits and a lack of standardized metrics during evaluation, which impedes reliable comparative analysis across different model architectures. Furthermore,  the data distribution in MIMIC-CXR, commonly used in model training, fails to adequately test the models' ability to generalize to new, unseen distributions.
To fill this gap, we introduce \textbf{ReXrank}~(\textbf{\href{https://rexrank.ai}{https://rexrank.ai}}), a public leaderboard and challenge specifically designed for evaluating AI-powered radiology report generation from chest X-ray images. 

ReXrank offers a comprehensive evaluation framework that sets a standardized benchmark for assessing the effectiveness of different radiology report generation models. To ensure robust and clinically relevant evaluations, it integrates diverse datasets, including MIMIC-CXR~\cite{johnson2019mimic}, IU-Xray~\cite{demner2016preparing}, CheXpert Plus~\cite{chambon2024chexpert}, and ReXGradient, a large-scale private dataset of 10,000 studies. This broad dataset spectrum allows us to evaluate model performance on data with varying distributions, providing deeper insights into the models' generalization capabilities.
Furthermore, ReXrank implements various report evaluation metrics, including BLEU-2~\cite{papineni2002bleu}, BERTScore~\cite{zhang2019bertscore}, SembScore~\cite{smit2020chexbert}, RadGraph-F1~\cite{yu2023evaluating}, RadCliQ~\cite{yu2023evaluating}, RaTEScore~\cite{zhao2024ratescore}, GREEN~\cite{ostmeier2024green}, FineRadScore~\cite{huang2024fineradscore}, etc., to offer a detailed view of each model's strengths and weaknesses. This comprehensive approach enables a more nuanced understanding of model performance and facilitates meaningful comparisons between different AI-powered radiology report generation systems.

\begin{figure*}[htb]
    \centering
    \includegraphics[width=1\linewidth]{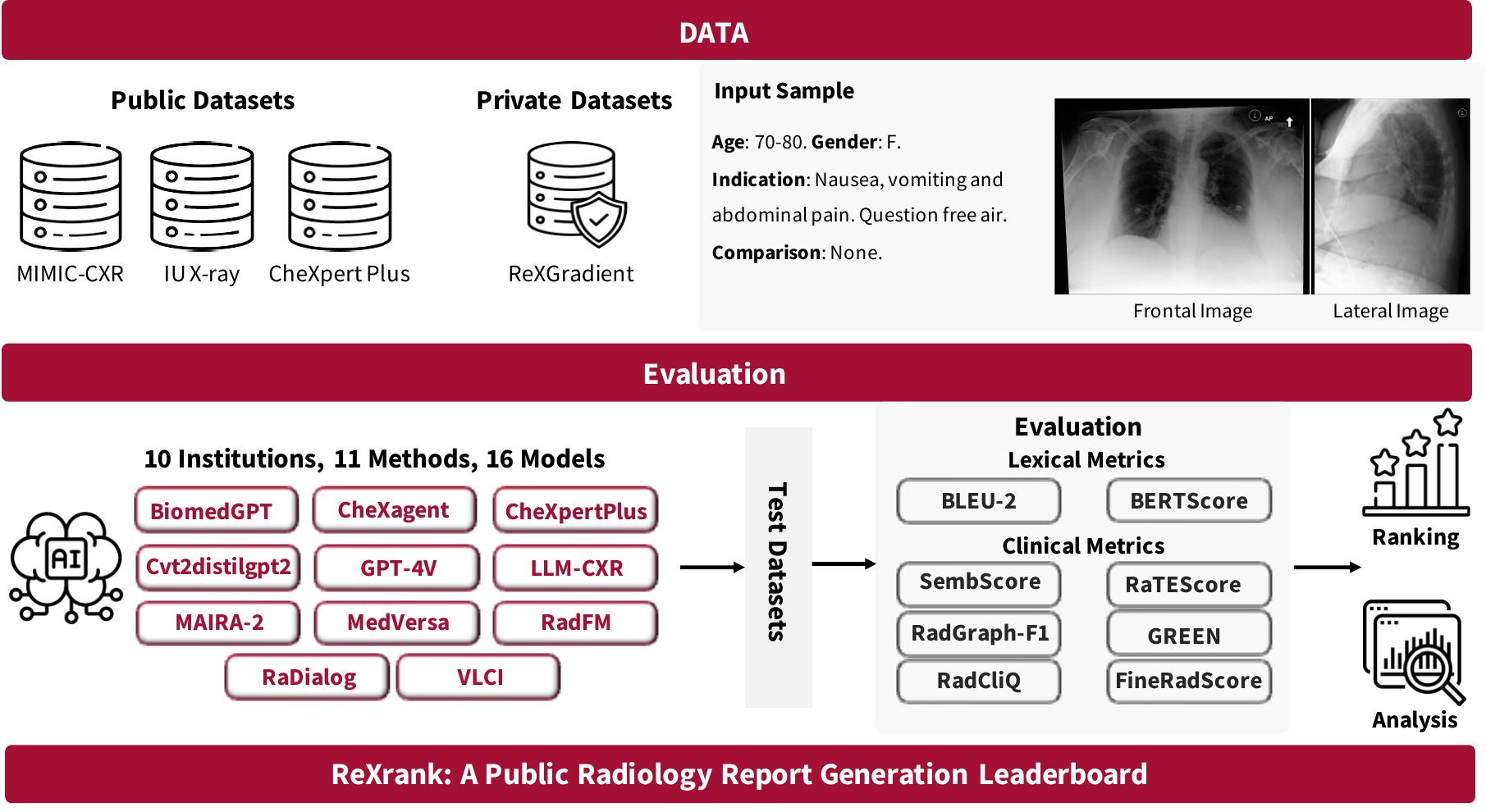}
    \caption{An illustration of ReXrank, a public leaderboard and challenge for AI-powered radiology report generation from chest X-ray images. ReXrank supports model submissions and evaluates them on both public datasets and a large-scale private dataset, providing comprehensive rankings of all submitted models.}
    \label{fig:teaser}
\end{figure*}

\section{Overview}
\noindent \textbf{Datasets.}
ReXrank leverages three public datasets and one comprehensive private dataset for report generation assessment. The private ReXGradient dataset comprises 10,000 studies from 67 U.S. medical sites, making it one of the largest and most geographically diverse evaluation sets. For public datasets, we utilize the official test splits of MIMIC-CXR (2,347 studies) and IU-Xray (590 studies), along with CheXpert Plus's validation set (200 studies) as no test split is available.

\vspace{3pt}
\noindent \textbf{Models.}
ReXrank currently includes \textbf{16} report generation models from 10 different institutions, including BiomedGPT\_IU~\cite{zhang2024generalist}, CheXagent~\cite{chen2024chexagent}, CheXpertPlus\_CheX~\cite{chambon2024chexpert}, CheXpertPlus\_CheX\_MIMIC~\cite{chambon2024chexpert}, CheXpertPlus\_MIMIC~\cite{chambon2024chexpert}, Cvt2distilgpt2\_IU~\cite{nicolson2023improving}, Cvt2distilgpt2\_MIMIC~\cite{nicolson2023improving}, GPT4V~\cite{yang2023dawn}, LLM-CXR~\cite{lee2023llm}, MAIRA-2~\cite{bannur2024maira}, MedVersa~\cite{zhou2024generalist}, RadFM~\cite{wu2023towards}, RaDialog~\cite{pellegrini2023radialog}, RGRG~\cite{tanida2023interactive}, VLCI\_IU~\cite{chen2023cross} and VLCI\_MIMIC~\cite{chen2023cross}. 
These models were trained on different medical datasets, primarily MIMIC-CXR, CheXpert Plus, and IU-Xray, with some models capable of handling multiple tasks beyond just report generation.

\vspace{3pt}
\noindent \textbf{Metrics.}
ReXrank employs \textbf{8} different metrics to comprehensively assess the quality of generated radiology reports, including traditional text generation metrics like BLEU-2~\cite{papineni2002bleu} and BERTScore~\cite{zhang2019bertscore}, as well as domain-specific metrics designed for radiology report evaluation such as SembScore~\cite{smit2020chexbert}, RadGraph-F1~\cite{yu2023evaluating}, RadCliQ-v1~\cite{yu2023evaluating} and RaTEScore~\cite{zhao2024ratescore}. The framework also incorporates recently developed LLM-based metrics including GREEN~\cite{ostmeier2024green}, and FineRadScore~\cite{huang2024fineradscore}, which focus on identifying clinically significant errors. Each metric evaluates different aspects of the generated reports, from textual similarity to clinical accuracy, providing a thorough assessment of model performance. 
We default use RadCliQ-v1 as the primary metric.

\begin{figure*}[!h]
\centering
\includegraphics[width=1\linewidth]{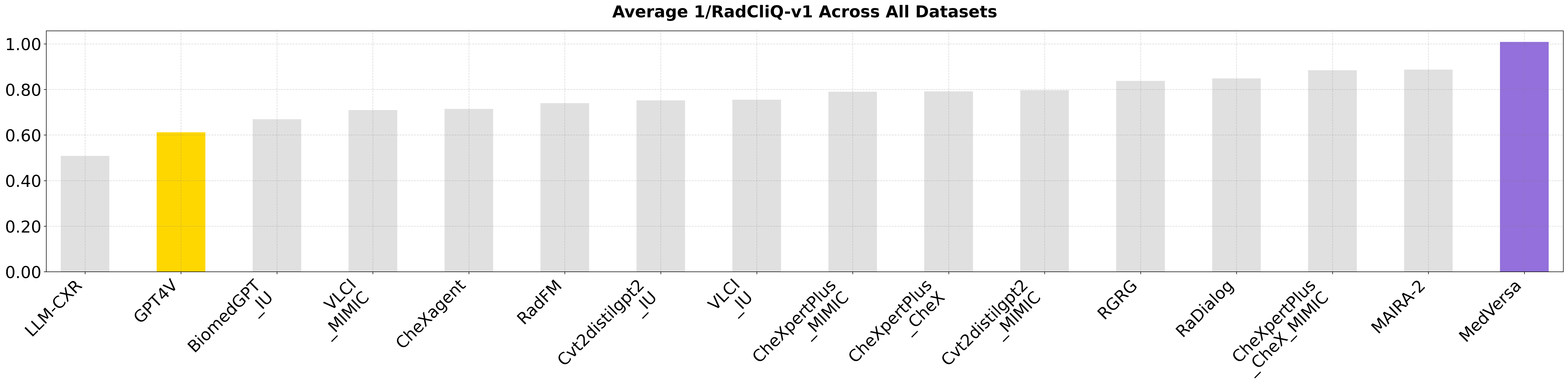}
\caption{Comprehensive performance evaluation and ranking of report generation models based on the average  \textbf{1/RadCliQ-v1} metric of four distinct datasets: ReXGradient, MIMIC-CXR, IU X-ray, and CheXpert Plus. MedVersa (highlighted in purple) demonstrates consistently superior performance, achieving significantly higher scores compared to other models, including GPT4V (highlighted in yellow).}
\label{fig:model_ranking}
\end{figure*}

\noindent \textbf{Results.}
MedVersa emerges as one of the top-performing models (Figure~\ref{fig:model_ranking}), with best 1/RadCliQ-v1 scores of 0.98 ± 0.05 on ReXGradient and 0.92 ± 0.02 on MIMIC-CXR. However, its performance on the CheXpert Plus dataset is comparatively lower, ranking fourth with a 1/RadCliQ-v1 score of 0.72 ± 0.10 on the Findings. 
MedVersa consistently outperforms GPT4V, the state-of-the-art generalist vision-language model, across multiple metrics and datasets.
We further analyze the distribution of evaluation metrics across datasets. Among the four datasets, IU X-ray stands out as the least challenging, consistently yielding high performance across models. In contrast, CheXpert Plus exhibits the highest variance and lower overall performance, likely due to its distinct data distribution and small validation set (200 studies). 
The private ReXGradient dataset demonstrates remarkably low-performance variance across models, underscoring its high data quality and utility as a benchmark for assessing model robustness.

\begin{figure*}[!h]
\centering
\includegraphics[width=1\linewidth]{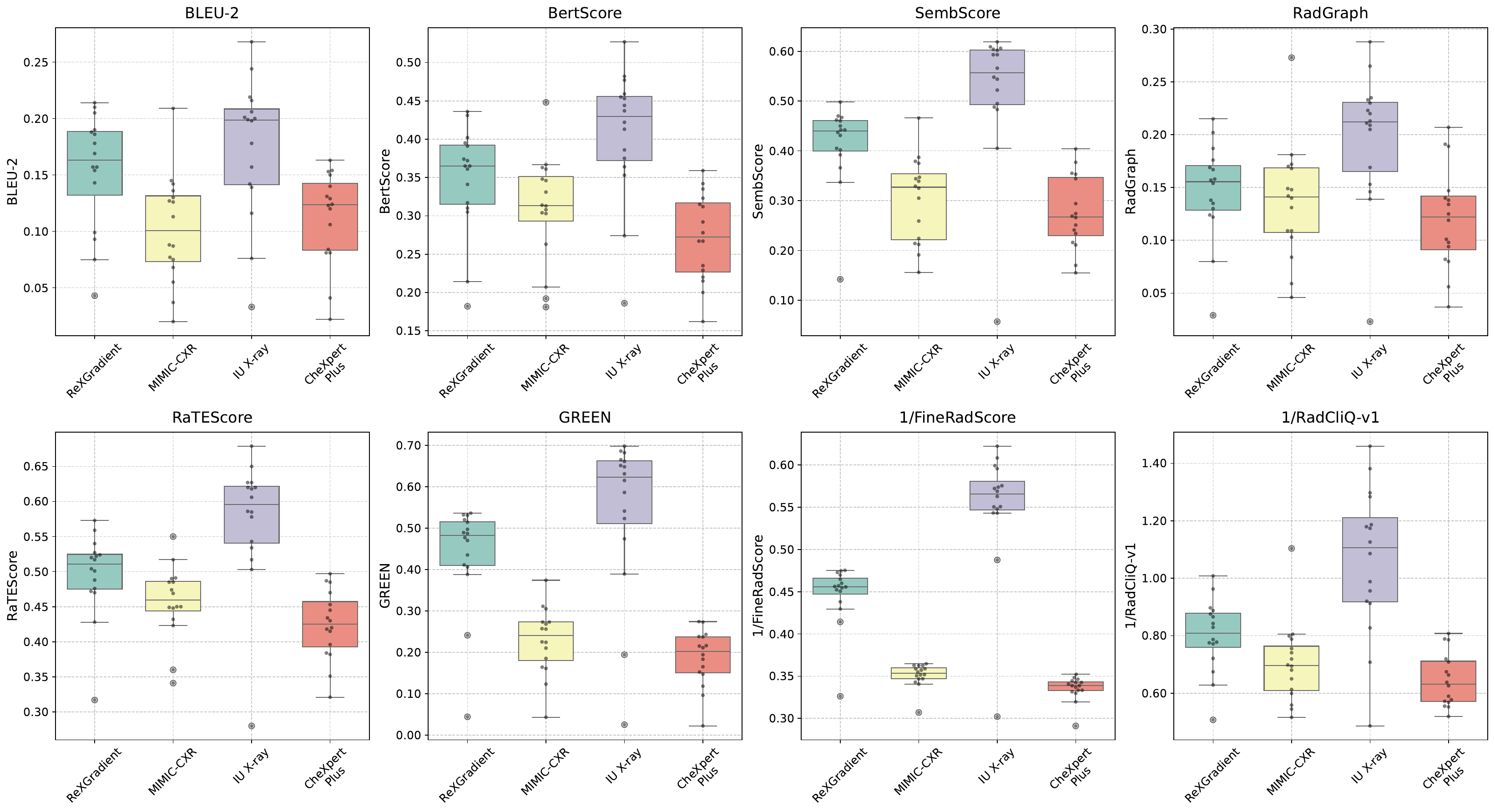}
\caption{Distribution of evaluation metrics across different datasets: ReXGradient, and MIMIC-CXR, IU X-ray, CheXpert Plus. Box plots show the variation in model performance for each metric. For consistency in visualization, we plot the reciprocals (1/x) of FineRadScore and RadCliQ-v1, so higher values indicate better performance across all metrics.}
\label{fig:metric_dataset}
\end{figure*}

\section{Method}

\subsection{Datasets}
Our evaluation leverages four distinct datasets: ReXGradient, MIMIC-CXR, IU-Xray, and CheXpert Plus. These datasets provide diverse testing distributions across different medical institutions and patient populations.

\textbf{ReXGradient.} This private test set is provided by Gradient Health, which consists of 10,000 studies collected from 7,004 patients across 67 medical sites in the United States. 

\textbf{MIMIC-CXR~\cite{johnson2019mimic}.} This is a large, publicly accessible dataset comprising 377,110 chest X-rays (CXRs) corresponding to 227,835 radiographic studies performed at the Beth Israel Deaconess Medical Center in Boston, MA. 
For this dataset, we extracted sections of indication, comparison, findings, and impression via keyword matching.
In our experiments, we follow MIMIC-CXR's official split and report scores on the test set, which consists of 2,347 studies.

\textbf{IU-Xray ~\cite{demner2016preparing}.} This is a publicly accessible dataset containing 7,470 pairs of CXRs and radiology reports. Each study in this dataset includes one frontal and one lateral CXR, associated with a single radiology report. We follow the split provided by R2Gen \cite{chen2020generating} and report scores on the test set, which comprises 590 studies.

\textbf{CheXpert Plus~\cite{chambon2024chexpert}.} This is a large, publicly accessible consisting of 223,462 unique pairs of radiology reports and chest X-rays.
These correspond to 187,711 radiographic studies from 64,725 patients. 
We follow the official split of CheXpert Plus and report scores on the validation set, which contains 200 studies.

\subsection{Data format}
For each study in our test datasets, the data is organized in a structured format. 
\begin{itemize}
\item \texttt{id}: Unique identifier for the study
\item \texttt{image\_path}: List of paths to all relevant chest X-ray images
\item \texttt{frontal\_lateral}: List indicating the view type of each image
\item \texttt{key\_image\_path}: Path to the primary image (typically frontal view)
\item \texttt{context}: Patient information and clinical context
\item \texttt{report}: Radiologist's findings and impressions
\end{itemize}

\subsection{Metrics}
\textbf{BLEU-2~\cite{papineni2002bleu}.} BLEU (Bilingual Evaluation Understudy) is a widely used metric in machine translation and text generation tasks. It evaluates the quality of generated text by comparing n-gram precision between the candidate and reference texts, with scores ranging from 0 to 1. In this work, we specifically use BLEU-2, which focuses on bigram precision to assess the quality of the generated text.

\textbf{BERTScore~\cite{zhang2019bertscore}.}  BERTScore is a neural metric that uses pre-trained BERT models~\cite{kenton2019bert} to evaluate text similarity. It computes cosine similarity between the BERT embeddings of model-generated and groundtruth radiology reports.

\textbf{SembScore~\cite{smit2020chexbert}.} SembScore (CheXbert labeler vector similarity) is a domain-specific metric for radiology report evaluation. It computes the cosine similarity between the indicator vectors of 14 pathologies that the CheXbert automatic labeler extracts from model-generated and groundtruth radiology reports.

\textbf{RadGraph-F1~\cite{yu2023evaluating}.} RadGraph-F1 is a metric for radiology report evaluation. It computes the overlap in clinical entities and relations that RadGraph~\cite{jain2021radgraph} extracts from candidate and reference reports. 

\textbf{1/RadCliQ-v1~\cite{yu2023evaluating}.} RadCliQ is a composite metric designed for evaluating radiology report generation, combining BLEU, BERTScore, SembScore, and RadGraph-F1 to provide a comprehensive assessment of generated reports. 
For our evaluation, we utilized the official implementation\footnote{\url{https://github.com/rajpurkarlab/CXR-Report-Metric}}, which also includes BLEU, BERTScore, SembScore, RadGraph-F1.
While the original RadCliQ metric is designed as lower-is-better, we first calculate the average RadCliQ-v1 score for each model across the dataset, then take its reciprocal (1/RadCliQ-v1) to maintain consistency with other metrics where higher values indicate better performance.

\textbf{RaTEScore~\cite{zhao2024ratescore}.} RaTEScore is an entity-aware metric for radiology report evaluation. It emphasizes crucial medical entities like diagnostic outcomes and anatomical details and is robust against complex medical synonyms and sensitive to negation expressions. For our evaluation, we utilized the official implementation\footnote{\url{https://github.com/MAGIC-AI4Med/RaTEScore}}.

\textbf{GREEN~\cite{ostmeier2024green}.} GREEN (Generative Radiology Report Evaluation and Error Notation) is an LLM-based metric for evaluating radiology report generation.
It leverages language models to identify and explain clinically significant errors in quantitative and qualitative candidate reports. We utilized the official implementation\footnote{\url{https://github.com/Stanford-AIMI/GREEN}} for our evaluation.

\textbf{1/FineRadScore~\cite{huang2024fineradscore}.} FineRadScore is an automated evaluation metric leveraging an LLM  to assess the quality of generated chest X-ray reports. It determines the minimum number of line-by-line corrections needed, with severity ratings from 1 to 4, to transform a candidate report into a ground-truth report. For evaluation, we employ GPT-4o as the LLM  and use the official implementation\footnote{\url{https://github.com/rajpurkarlab/FineRadScore}}. To obtain a single FineRadScore for each report, we take the maximum clinical severity across all lines.
Similar to 1/RadCliQ-v1, we present 1/FineRadScore to maintain consistency where higher values indicate better performance.

\subsection{Confidence Intervals}
In our analysis, we generate confidence intervals (CIs) by assuming a normal distribution of data. This statistical method calculates the mean and standard deviation of our data and then uses the standard error of the mean to estimate variability. 
For a 95\% confidence level, a Z-score of approximately 1.96 is used to determine the interval. 
This Z-score indicates that the true mean is likely within 1.96 standard errors of the sample mean. 
By multiplying the Z-score by the standard error, we obtain the CI, providing a range that encapsulates the true average value with 95\% certainty. 

\subsection{Participating Models}

\textbf{BiomedGPT\_IU~\cite{zhang2024generalist}.} BiomedGPT is a lightweight, open-source vision-language model designed for diverse biomedical tasks across modalities. The model was fine-tuned for VQA and image captioning tasks using multiple datasets, including radiology and pathology data. BiomedGPT\_IU is fine-tuned on the IU X-ray dataset for image captioning tasks. In our evaluation, we used the publicly available checkpoints trained on the IU-Xray dataset\footnote{\url{https://github.com/taokz/BiomedGPT}}.


\textbf{CheXagent~\cite{chen2024chexagent}.} CheXagent is an instruction-tuned foundation model specifically designed for chest X-ray interpretation. The model consists of a vision encoder for representing CXR images, and a network to bridge the vision and language modalities. This model is trained on CheXinstruct, a large-scale instruction-tuning dataset curated from 28 publicly-available datasets.
For our evaluation, we utilized the publicly available 8 billion parameter checkpoint from Hugging Face\footnote{\url{https://huggingface.co/StanfordAIMI/CheXagent-8b}}.

\textbf{CheXpertPlus\_CheX~\cite{chambon2024chexpert}.} CheXpertPlus\_CheX, introduced in the CheXpert Plus paper, utilizes a Swinv2~\cite{liu2022swin} architecture with a two-layer BERT decoder~\cite{kenton2019bert} for medical report generation. CheXpertPlus\_CheX is trained exclusively on the CheXpert Plus dataset. 
In our evaluation, we utilized the publicly available Findings Checkpoint\footnote{\url{https://huggingface.co/IAMJBchexpert-findings-baseline}} and Impression Checkpoint\footnote{\url{https://huggingface.co/IAMJB/chexpert-impression-baseline}}.
Our evaluation employs these models sequentially, generating the findings and impression sections separately, and then combining them with appropriate headers to form the complete report.

\textbf{CheXpertPlus\_CheX\_MIMIC~\cite{chambon2024chexpert}.} CheXpertPlus\_CheX\_MIMIC shares the same architectural design as CheXpertPlus\_MIMIC, employing the Swinv2 architecture with a two-layer BERT decoder. CheXpertPlus\_CheX is trained exclusively on the combination of MIMIC-CXR and CheXpert Plus dataset. In our evaluation, we utilized the publicly available Findings Checkpoint\footnote{\url{https://huggingface.co/IAMJB/chexpert-mimic-cxr-findings-baseline}} and Impression Checkpoint\footnote{\url{https://huggingface.co/IAMJB/chexpert-mimic-cxr-impression-baseline}}.
Our evaluation employs these models sequentially, generating the findings and impression sections separately, and then combining them with appropriate headers to form the complete report.

\textbf{CheXpertPlus\_MIMIC~\cite{chambon2024chexpert}.} CheXpertPlus\_MIMIC shares the same architectural design as CheXpertPlus\_CheX, employing the Swinv2 architecture with a two-layer BERT decoder. CheXpertPlus\_MIMIC comprises two distinct models trained on MIMIC-CXR: one for findings and another for impressions. 
In our evaluation, we utilized the publicly available Findings Checkpoint\footnote{\url{https://huggingface.co/IAMJB/mimic-cxr-findings-baseline}} and Impression Checkpoint\footnote{\url{https://huggingface.co/IAMJB/mimic-cxr-impression-baseline}}.
Our evaluation employs these models sequentially, generating the findings and impression sections separately, and then combining them with appropriate headers to form the complete report.

\textbf{Cvt2distilgpt2\_IU~\cite{nicolson2023improving}} CvT2DistilGPT2\_IU employs a hybrid architecture combining a Convolutional vision Transformer (CvT) encoder~\cite{wu2021cvt} pre-trained on ImageNet-21K~\cite{deng2009imagenet} with a DistilGPT2~\cite{alfarghaly2021automated} decoder for chest X-ray report generation. This model leverages the CvT's efficient hierarchical design for image feature extraction and DistilGPT2's natural language generation capabilities.  In our evaluation, we utilized the publicly available checkpoint from Github\footnote{\url{https://github.com/aehrc/cvt2distilgpt2}} and followed the official evaluation guidelines. 

\textbf{Cvt2distilgpt2\_MIMIC~\cite{nicolson2023improving}} Cvt2distilgpt2\_MIMIC applies the same Cvt2distilgpt2 architecture but is trained on the MIMIC-CXR dataset. Our evaluation utilized the publicly available MIMIC-CXR-trained checkpoints.

\textbf{RGRG~\cite{tanida2023interactive}.} RGRG (Region-Guided Radiology Report Generation) employs object detection to extract localized visual features from 29 anatomical regions in chest X-rays. It uses binary classifiers to select salient features and encode abnormalities, followed by a language model generating sentences for each selected region. RGRG was trained on the Chest ImaGenome dataset~\cite{wu2021chest}. 
In our evaluation, we utilized the publicly available checkpoint from Github\footnote{\url{https://github.com/ttanida/rgrg}} and followed the official evaluation guidelines. 

\textbf{RaDialog~\cite{pellegrini2023radialog}.}  RaDialog is a large vision-language model for radiology report generation and interactive dialogue. It integrates visual image features and structured pathology findings with a large language model (LLM), adapted to radiology using parameter-efficient fine-tuning. RaDialog was trained on the MIMIC-CXR for radiology report generation tasks. 
In our evaluation, we utilized the publicly available LLaVA version checkpoint from Huggingface\footnote{\url{https://huggingface.co/ChantalPellegrini/RaDialog-interactive-radiology-report-generation}} and followed the official evaluation guidelines.

\textbf{GPT-4V~\cite{yang2023dawn}.} GPT-4V (GPT-4 with vision) is a multimodal large language model released by OpenAI, which enables users to instruct GPT-4 to analyze image inputs provided by the user. In our evaluation, we used the API of model ``gpt4o05132024'' and followed the official evaluation protocols to assess its performance. The prompt we used is ``You are a helpful assistant. Please generate a report for the given images, including both findings and impressions. Return the report in the following format: Findings: \{\} Impression: \{\}. ''.

\textbf{LLM-CXR~\cite{lee2023llm}.} LLM-CXR is a multimodal large language model that utilizes VQ-GAN to tokenize images, integrating both image and text tokens as input to its base LLM architecture. This model enables CXR-to-report generation, report-to-CXR generation, and CXR-related visual question answering (VQA). For our evaluation, we used the publicly available checkpoints\footnote{\url{https://github.com/hyn2028/llm-cxr}} and followed the official evaluation guidelines.

\textbf{MAIRA-2~\cite{bannur2024maira}.} MAIRA-2 is a multimodal large language model that combines a radiology-specific image encoder with a Large Language Model (LLM), trained for grounded report generation from chest X-rays. For input, the model accepts X-ray images along with indication, comparison, and technique information. For our evaluation, we used the publicly available checkpoints\footnote{\url{https://huggingface.co/microsoft/maira-2}} and followed the official evaluation guidelines.
For studies containing both frontal and lateral views, we input the technique that ``PA and lateral views of the chest were obtained.''. For studies with only frontal views, we use ``PA view of the chest was obtained.''.

\textbf{MedVersa~\cite{zhou2024generalist}.} MedVersa is a versatile medical AI system that can coordinate multiple models and tools to perform various tasks and generate multimodal outputs. 
MedVersa was trained on the MIMIC-CXR training and validation dataset for medical report generation tasks. In our evaluation, we utilized the publicly available checkpoint from Huggingface\footnote{\url{https://huggingface.co/hyzhou/MedVersa}} and followed the official evaluation guidelines. The standard prompt structure we employed for report generation was ``Can you provide a report of \{input\_image\_token\} with findings and impression?'', where \{input\_image\_token\} represents the placeholder for the input image.

\textbf{RadFM~\cite{wu2023towards}.} RadFM is a radiology foundation model trained on large-scale multi-modal datasets. It supports both 2D and 3D scans, multi-image input, and visual-language interleaving cases. The model's training included the MIMIC-CXR dataset. For our evaluation, we utilized the publicly available checkpoint from Huggingface\footnote{\url{https://huggingface.co/chaoyi-wu/RadFM}} and followed the official evaluation guidelines. The prompt we employed for report generation was ``Can you provide a radiology report for this medical image?''.

\textbf{VLCI\_IU~\cite{chen2023cross}.} VLCI~(Visual-Linguistic Causal Intervention) is a vision-language model using a multiway transformer for cross-modal alignment with Visual-linguistic causal intervention, integrating a pre-trained transformer and Visual and linguistic de-confounding Modules to mitigate cross-modal bias through local and global visual sampling and linguistic estimation using a vocabulary dictionary and visual features.
In our evaluation, we used the publicly available checkpoints trained on the IU-Xray dataset\footnote{\url{https://github.com/WissingChen/VLCI}}.

\textbf{VLCI\_MIMIC~\cite{chen2023cross}.} VLCI\_MIMIC applies the same VLCI architecture but is trained on the MIMIC-CXR dataset. Our evaluation utilized the publicly available MIMIC-CXR-trained checkpoints.
\section{Results}

Table~\ref{tab:private_leaderboard},~\ref{tab:mimiccxr_leaderboard},~\ref{tab:iuxray_leaderboard} and ~\ref{tab:chexpert_leaderboard} summarize the performance of various medical report generation models across four different datasets: ReXGradient, MIMIC-CXR, IU X-ray and CheXpert Plus.
Among these, MedVersa demonstrates superior performance, achieving the best 1/RadCliQ-v1 scores on ReXGradient (1.01 ± 0.01), MIMIC-CXR (1.10 ± 0.02), and IU X-ray (1.46 ± 0.03) on the findings section. 
This consistent top performance across different datasets indicates MedVersa's robust generalization capabilities and effectiveness in medical report generation.

Moreover, the results displayed in the tables show that ReXGradient serves as a dataset where models consistently exhibit minimal confidence intervals of 0.01 for most models on the RadCliQ-v1 metric, thereby supporting its utility as a reliable benchmark for medical report generation models.
Figure~\ref{fig:metric_dataset} illustrates the distribution of evaluation metrics.
The IU X-ray dataset, while always showing high-performance scores (the best model achieving a 1/RadCliQ-v1 score of 1.46 ± 0.03 on the findings sections), suggests that it may be too simplistic or lacking in complexity necessary for rigorous model differentiation.
In contrast, CheXpert Plus shows lower overall performance (the best model obtaining a 1/RadCliQ-v1 score of 0.81 ± 0.12 on the findings sections) with higher variance, potentially indicating dataset distribution shifts or noise.

Models trained on multiple datasets (e.g., CheXpertPlus\_CheX\_MIMIC) tend to outperform those trained on individual datasets, suggesting that a multi-dataset training approach helps bridge the distributional gap and enhances generalization. 
Models perform better when evaluated on the same distribution seen in training, for instance, VLCI\_IU achieves superior performance on IU X-ray (RadCliQ-v1: 1.38 ± 0.04) compared to VLCI\_MIMIC (0.91 ± 0.04), while VLCI\_MIMIC performs better on MIMIC-CXR (0.68 ± 0.02 vs 0.60 ± 0.02 for VLCI\_IU).

The comparison between findings-only and findings + impression tasks reveals interesting model behaviors. On ReXGradient, MedVersa shows slight performance degradation when generating both findings and impressions (1/RadCliQ-v1 decreasing from 1.01 ± 0.01 to 0.98 ± 0.05), while CheXpertPlus\_CheX\_MIMIC shows improvement (from 0.83 ± 0.01 to 0.85 ± 0.01). 
This may be due to CheXpertPlus\_CheX\_MIMIC using separate models for findings and impression generation, while MedVersa only uses a single model architecture. The separate models may allow CheXpertPlus\_CheX\_MIMIC to better specialize in each subtask.


\begin{table}[!h]
\centering
\caption{Comprehensive evaluation of medical report generation models on ReXGradient. Models are ranked by 1/RadCliQ-v1.  Model evaluation results with 95\% confidence intervals (mean ± CI) under normality assumption. The best results for each metric are shown in \textbf{bold}.}
\label{tab:private_leaderboard}
\vspace{6pt}

\renewcommand{\arraystretch}{1.2}
\setlength{\tabcolsep}{3pt}

\resizebox{\textwidth}{!}{
\begin{tabular}{lcccccccc}
\toprule
\rowcolor{gray!10}
\textbf{Model} & 
\textbf{1/RadCliQ-v1} $\uparrow$ &
\textbf{BLEU-2} $\uparrow$ & 
\textbf{BertScore} $\uparrow$ & 
\textbf{SembScore} $\uparrow$ & 
\textbf{RadGraph} $\uparrow$ & 
\textbf{RaTEScore} $\uparrow$ & 
\textbf{GREEN} $\uparrow$ & 
\textbf{1/FineRadScore} $\uparrow$ \\
\midrule
\multicolumn{9}{l}{\textbf{Findings + Impression}} \\
\midrule
MedVersa & \textbf{0.98 ± 0.01} & 0.17 ± 0.01 & \textbf{0.44 ± 0.01} & \textbf{0.48 ± 0.02} & \textbf{0.19 ± 0.01} & \textbf{0.53 ± 0.00} & \textbf{0.52 ± 0.01} & \textbf{0.47 ± 0.02} \\
CheXpertPlus\_CheX\_MIMIC & 0.85 ± 0.01 & \textbf{0.20 ± 0.00} & 0.39 ± 0.00 & 0.43 ± 0.00 & 0.17 ± 0.00 & 0.50 ± 0.00 & 0.51 ± 0.01 & 0.47 ± 0.02 \\
CheXpertPlus\_MIMIC & 0.80 ± 0.01 & 0.18 ± 0.00 & 0.36 ± 0.00 & 0.43 ± 0.00 & 0.14 ± 0.00 & 0.48 ± 0.00 & 0.52 ± 0.01 & 0.47 ± 0.02 \\
CheXpertPlus\_CheX & 0.76 ± 0.01 & 0.17 ± 0.00 & 0.33 ± 0.00 & 0.40 ± 0.00 & 0.15 ± 0.00 & 0.50 ± 0.00 & 0.47 ± 0.01 & 0.42 ± 0.02 \\
RadFM & 0.74 ± 0.01 & 0.13 ± 0.00 & 0.34 ± 0.00 & 0.38 ± 0.00 & 0.13 ± 0.00 & 0.47 ± 0.00 & 0.41 ± 0.01 & 0.43 ± 0.02 \\
GPT4V & 0.66 ± 0.01 & 0.07 ± 0.00 & 0.21 ± 0.00 & 0.36 ± 0.00 & 0.17 ± 0.00 & 0.46 ± 0.00 & 0.36 ± 0.01 & 0.42 ± 0.02 \\
\midrule
\multicolumn{9}{l}{\textbf{Findings}} \\
\midrule
MedVersa & \textbf{1.01 ± 0.01} & 0.21 ± 0.00 & 0.43 ± 0.00 & \textbf{0.50 ± 0.00} & 0.20 ± 0.00 & 0.53 ± 0.00 & 0.53 ± 0.01 & \textbf{0.47 ± 0.02} \\
MAIRA-2 & 0.96 ± 0.01 & 0.20 ± 0.00 & \textbf{0.44 ± 0.00} & 0.46 ± 0.00 & 0.19 ± 0.00 & 0.56 ± 0.00 & 0.53 ± 0.01 & \textbf{0.47 ± 0.02} \\
VLCI\_IU & 0.90 ± 0.01 & \textbf{0.21 ± 0.00} & 0.36 ± 0.00 & 0.47 ± 0.00 & \textbf{0.21 ± 0.00} & \textbf{0.57 ± 0.00} & \textbf{0.54 ± 0.01} & 0.45 ± 0.02 \\
RGRG & 0.89 ± 0.01 & 0.19 ± 0.00 & 0.39 ± 0.00 & 0.47 ± 0.00 & 0.17 ± 0.00 & 0.54 ± 0.00 & 0.49 ± 0.01 & 0.46 ± 0.02 \\
RaDialog & 0.88 ± 0.01 & 0.19 ± 0.00 & 0.40 ± 0.00 & 0.45 ± 0.00 & 0.16 ± 0.00 & 0.52 ± 0.00 & 0.43 ± 0.01 & 0.46 ± 0.02 \\
Cvt2distilgpt2\_MIMIC & 0.87 ± 0.01 & 0.19 ± 0.00 & 0.37 ± 0.00 & 0.46 ± 0.00 & 0.18 ± 0.00 & 0.52 ± 0.00 & 0.51 ± 0.01 & 0.47 ± 0.02 \\
Cvt2distilgpt2\_IU & 0.84 ± 0.01 & 0.18 ± 0.00 & 0.40 ± 0.00 & 0.41 ± 0.00 & 0.17 ± 0.00 & 0.52 ± 0.00 & 0.47 ± 0.01 & 0.46 ± 0.02 \\
CheXpertPlus\_CheX\_MIMIC & 0.83 ± 0.01 & 0.17 ± 0.00 & 0.37 ± 0.00 & 0.44 ± 0.00 & 0.15 ± 0.00 & 0.52 ± 0.00 & 0.49 ± 0.01 & 0.47 ± 0.02 \\
CheXpertPlus\_CheX & 0.79 ± 0.01 & 0.14 ± 0.00 & 0.36 ± 0.00 & 0.43 ± 0.00 & 0.12 ± 0.00 & 0.48 ± 0.00 & 0.41 ± 0.01 & 0.41 ± 0.02 \\
CheXpertPlus\_MIMIC & 0.78 ± 0.01 & 0.15 ± 0.00 & 0.34 ± 0.00 & 0.44 ± 0.00 & 0.13 ± 0.00 & 0.50 ± 0.00 & 0.52 ± 0.01 & 0.47 ± 0.02 \\
RadFM & 0.78 ± 0.01 & 0.16 ± 0.00 & 0.36 ± 0.00 & 0.39 ± 0.00 & 0.14 ± 0.00 & 0.50 ± 0.00 & 0.41 ± 0.01 & 0.44 ± 0.02 \\
BiomedGPT\_IU & 0.77 ± 0.01 & 0.10 ± 0.00 & 0.32 ± 0.00 & 0.44 ± 0.00 & 0.16 ± 0.00 & 0.47 ± 0.00 & 0.39 ± 0.01 & 0.45 ± 0.02 \\
VLCI\_MIMIC & 0.72 ± 0.01 & 0.16 ± 0.00 & 0.31 ± 0.00 & 0.40 ± 0.00 & 0.12 ± 0.00 & 0.49 ± 0.00 & 0.48 ± 0.01 & 0.46 ± 0.02 \\
CheXagent & 0.67 ± 0.01 & 0.09 ± 0.00 & 0.30 ± 0.00 & 0.37 ± 0.00 & 0.08 ± 0.00 & 0.43 ± 0.00 & 0.24 ± 0.01 & 0.46 ± 0.02 \\
GPT4V & 0.63 ± 0.01 & 0.07 ± 0.00 & 0.21 ± 0.00 & 0.34 ± 0.00 & 0.14 ± 0.00 & 0.47 ± 0.00 & 0.50 ± 0.01 & 0.43 ± 0.02 \\
LLM-CXR & 0.51 ± 0.01 & 0.04 ± 0.00 & 0.18 ± 0.00 & 0.14 ± 0.00 & 0.03 ± 0.00 & 0.32 ± 0.00 & 0.04 ± 0.00 & 0.33 ± 0.02 \\
\bottomrule
\end{tabular}
}

\end{table}

\begin{table}[!h]
\centering
\caption{Comprehensive evaluation of medical report generation models on the MIMIC-CXR datasets. Models are ranked by 1/RadCliQ-v1.  Model evaluation results with 95\% confidence intervals (mean ± CI) under normality assumption. The best results for each metric are shown in \textbf{bold}.}
\label{tab:mimiccxr_leaderboard}
\vspace{6pt}
\renewcommand{\arraystretch}{1.2}
\setlength{\tabcolsep}{3pt}

\resizebox{\textwidth}{!}{
\begin{tabular}{lcccccccc}
\toprule
\rowcolor{gray!10}
\textbf{Model} & 
\textbf{1/RadCliQ-v1} $\uparrow$ &
\textbf{BLEU-2} $\uparrow$ & 
\textbf{BertScore} $\uparrow$ & 
\textbf{SembScore} $\uparrow$ & 
\textbf{RadGraph} $\uparrow$ & 
\textbf{RaTEScore} $\uparrow$ & 
\textbf{GREEN} $\uparrow$ & 
\textbf{1/FineRadScore} $\uparrow$ \\
\midrule
\multicolumn{9}{l}{\textbf{Findings + Impression}} \\
\midrule
MedVersa & \textbf{0.92 ± 0.02} & \textbf{0.19 ± 0.00} & \textbf{0.43 ± 0.01} & 0.32 ± 0.01 & \textbf{0.27 ± 0.01} & \textbf{0.55 ± 0.01} & \textbf{0.42 ± 0.01} & 0.36 ± 0.03 \\
CheXpertPlus\_CheX\_MIMIC & 0.83 ± 0.02 & 0.17 ± 0.00 & 0.36 ± 0.00 & \textbf{0.39 ± 0.01} & 0.20 ± 0.00 & 0.52 ± 0.00 & 0.37 ± 0.01 & \textbf{0.36 ± 0.03} \\
CheXpertPlus\_MIMIC & 0.80 ± 0.02 & 0.17 ± 0.00 & 0.35 ± 0.00 & 0.38 ± 0.01 & 0.19 ± 0.00 & 0.51 ± 0.00 & 0.38 ± 0.01 & \textbf{0.36 ± 0.03} \\
CheXpertPlus\_CheX & 0.71 ± 0.01 & 0.13 ± 0.00 & 0.30 ± 0.00 & 0.34 ± 0.01 & 0.17 ± 0.00 & 0.51 ± 0.00 & 0.30 ± 0.01 & 0.35 ± 0.03 \\
RadFM & 0.62 ± 0.02 & 0.08 ± 0.00 & 0.28 ± 0.00 & 0.24 ± 0.01 & 0.11 ± 0.00 & 0.45 ± 0.00 & 0.21 ± 0.01 & 0.35 ± 0.03 \\
GPT4V & 0.55 ± 0.01 & 0.07 ± 0.00 & 0.20 ± 0.00 & 0.19 ± 0.01 & 0.09 ± 0.00 & 0.43 ± 0.00 & 0.13 ± 0.01 & 0.33 ± 0.03 \\
\midrule
\multicolumn{9}{l}{\textbf{Findings}} \\
\midrule
MedVersa & \textbf{1.10 ± 0.02} & \textbf{0.21 ± 0.00} & \textbf{0.45 ± 0.01} & \textbf{0.47 ± 0.01} & \textbf{0.27 ± 0.01} & \textbf{0.55 ± 0.01} & \textbf{0.37 ± 0.01} & \textbf{0.36 ± 0.03} \\
CheXpertPlus\_CheX\_MIMIC & 0.81 ± 0.02 & 0.14 ± 0.00 & 0.37 ± 0.00 & 0.38 ± 0.01 & 0.18 ± 0.01 & 0.49 ± 0.01 & 0.30 ± 0.01 & 0.36 ± 0.03 \\
RaDialog & 0.80 ± 0.02 & 0.13 ± 0.00 & 0.36 ± 0.00 & 0.39 ± 0.01 & 0.17 ± 0.00 & 0.48 ± 0.00 & 0.27 ± 0.01 & 0.36 ± 0.03 \\
CheXpertPlus\_MIMIC & 0.79 ± 0.02 & 0.14 ± 0.00 & 0.36 ± 0.00 & 0.38 ± 0.01 & 0.17 ± 0.00 & 0.48 ± 0.01 & 0.31 ± 0.01 & 0.36 ± 0.03 \\
RGRG & 0.76 ± 0.02 & 0.13 ± 0.00 & 0.35 ± 0.00 & 0.34 ± 0.01 & 0.17 ± 0.00 & 0.49 ± 0.00 & 0.27 ± 0.01 & 0.35 ± 0.03 \\
CheXagent & 0.74 ± 0.02 & 0.11 ± 0.00 & 0.35 ± 0.01 & 0.35 ± 0.01 & 0.15 ± 0.00 & 0.47 ± 0.01 & 0.26 ± 0.01 & 0.35 ± 0.03 \\
Cvt2distilgpt2\_MIMIC & 0.72 ± 0.02 & 0.13 ± 0.00 & 0.33 ± 0.01 & 0.33 ± 0.01 & 0.15 ± 0.01 & 0.43 ± 0.01 & 0.27 ± 0.01 & 0.36 ± 0.03 \\
CheXpertPlus\_CheX & 0.70 ± 0.01 & 0.08 ± 0.00 & 0.31 ± 0.00 & 0.33 ± 0.01 & 0.14 ± 0.00 & 0.47 ± 0.00 & 0.23 ± 0.01 & 0.35 ± 0.03 \\
MAIRA-2 & 0.69 ± 0.02 & 0.09 ± 0.00 & 0.31 ± 0.00 & 0.34 ± 0.01 & 0.13 ± 0.00 & 0.52 ± 0.00 & 0.22 ± 0.01 & 0.36 ± 0.03 \\
VLCI\_MIMIC & 0.68 ± 0.02 & 0.14 ± 0.00 & 0.30 ± 0.00 & 0.30 ± 0.01 & 0.14 ± 0.00 & 0.45 ± 0.01 & 0.26 ± 0.01 & 0.36 ± 0.03 \\
RadFM & 0.65 ± 0.02 & 0.09 ± 0.00 & 0.31 ± 0.00 & 0.26 ± 0.01 & 0.11 ± 0.00 & 0.45 ± 0.01 & 0.18 ± 0.01 & 0.35 ± 0.03 \\
Cvt2distilgpt2\_IU & 0.61 ± 0.02 & 0.06 ± 0.00 & 0.30 ± 0.00 & 0.19 ± 0.01 & 0.10 ± 0.00 & 0.45 ± 0.01 & 0.16 ± 0.01 & 0.35 ± 0.03 \\
VLCI\_IU & 0.60 ± 0.02 & 0.07 ± 0.00 & 0.26 ± 0.00 & 0.21 ± 0.01 & 0.11 ± 0.00 & 0.45 ± 0.01 & 0.21 ± 0.01 & 0.35 ± 0.03 \\
GPT4V & 0.56 ± 0.01 & 0.07 ± 0.00 & 0.21 ± 0.00 & 0.21 ± 0.01 & 0.08 ± 0.00 & 0.42 ± 0.00 & 0.16 ± 0.01 & 0.34 ± 0.03 \\
BiomedGPT\_IU & 0.54 ± 0.01 & 0.02 ± 0.00 & 0.19 ± 0.00 & 0.22 ± 0.01 & 0.06 ± 0.00 & 0.36 ± 0.00 & 0.12 ± 0.01 & 0.34 ± 0.03 \\
LLM-CXR & 0.52 ± 0.01 & 0.04 ± 0.00 & 0.18 ± 0.00 & 0.16 ± 0.01 & 0.05 ± 0.00 & 0.34 ± 0.00 & 0.04 ± 0.00 & 0.31 ± 0.03 \\
\bottomrule
\end{tabular}
}

\end{table}

\begin{table}[!h]
\centering
\caption{Comprehensive evaluation of medical report generation models on the IU X-ray datasets. Models are ranked by 1/RadCliQ-v1.  Model evaluation results with 95\% confidence intervals (mean ± CI) under normality assumption. The best results for each metric are shown in \textbf{bold}.}
\label{tab:iuxray_leaderboard}
\vspace{6pt}

\renewcommand{\arraystretch}{1.2}
\setlength{\tabcolsep}{3pt}

\resizebox{\textwidth}{!}{
\begin{tabular}{lcccccccc}
\toprule
\rowcolor{gray!10}
\textbf{Model} & 
\textbf{1/RadCliQ-v1} $\uparrow$ &
\textbf{BLEU} $\uparrow$ & 
\textbf{BertScore} $\uparrow$ & 
\textbf{SembScore} $\uparrow$ & 
\textbf{RadGraph} $\uparrow$ & 
\textbf{RaTEScore} $\uparrow$ & 
\textbf{GREEN} $\uparrow$ & 
\textbf{1/FineRadScore} $\uparrow$ \\
\midrule
\multicolumn{9}{l}{\textbf{Findings + Impression}} \\
\midrule
MedVersa & \textbf{1.45 ± 0.04} & 0.20 ± 0.01 & \textbf{0.52 ± 0.01} & \textbf{0.60 ± 0.02} & \textbf{0.24 ± 0.01} & \textbf{0.63 ± 0.01} & 0.66 ± 0.02 & 0.58 ± 0.07 \\
CheXpertPlus\_CheX\_MIMIC & 1.28 ± 0.04 & \textbf{0.24 ± 0.01} & 0.48 ± 0.01 & 0.60 ± 0.02 & 0.23 ± 0.01 & 0.61 ± 0.01 & 0.69 ± 0.02 & 0.59 ± 0.07 \\
RadFM & 1.23 ± 0.05 & 0.20 ± 0.01 & 0.48 ± 0.01 & 0.56 ± 0.02 & 0.23 ± 0.01 & 0.60 ± 0.01 & 0.64 ± 0.02 & 0.55 ± 0.08 \\
CheXpertPlus\_MIMIC & 1.13 ± 0.04 & 0.23 ± 0.01 & 0.45 ± 0.01 & 0.59 ± 0.02 & 0.19 ± 0.01 & 0.57 ± 0.01 & 0.68 ± 0.02 & \textbf{0.61 ± 0.07} \\
CheXpertPlus\_CheX & 1.01 ± 0.03 & 0.20 ± 0.01 & 0.39 ± 0.01 & 0.55 ± 0.02 & 0.21 ± 0.01 & 0.60 ± 0.01 & \textbf{0.71 ± 0.02} & 0.57 ± 0.07 \\
GPT4V & 0.68 ± 0.03 & 0.08 ± 0.00 & 0.23 ± 0.01 & 0.40 ± 0.02 & 0.16 ± 0.01 & 0.52 ± 0.01 & 0.40 ± 0.03 & 0.53 ± 0.08 \\
\midrule
\multicolumn{9}{l}{\textbf{Findings}} \\
\midrule
MedVersa & \textbf{1.46 ± 0.03} & 0.21 ± 0.01 & \textbf{0.53 ± 0.01} & 0.61 ± 0.02 & 0.23 ± 0.01 & 0.65 ± 0.01 & 0.63 ± 0.02 & 0.57 ± 0.07 \\
VLCI\_IU & 1.38 ± 0.04 & \textbf{0.27 ± 0.01} & 0.46 ± 0.01 & \textbf{0.62 ± 0.02} & \textbf{0.29 ± 0.01} & \textbf{0.68 ± 0.01} & \textbf{0.70 ± 0.02} & 0.55 ± 0.07 \\
MAIRA-2 & 1.30 ± 0.04 & 0.22 ± 0.01 & 0.48 ± 0.01 & 0.60 ± 0.02 & 0.23 ± 0.01 & 0.63 ± 0.01 & 0.19 ± 0.02 & 0.60 ± 0.07 \\
Cvt2distilgpt2\_IU & 1.28 ± 0.05 & 0.24 ± 0.01 & 0.48 ± 0.01 & 0.55 ± 0.02 & 0.27 ± 0.02 & 0.62 ± 0.01 & 0.69 ± 0.02 & 0.56 ± 0.08 \\
RadFM & 1.19 ± 0.04 & 0.20 ± 0.01 & 0.46 ± 0.01 & 0.57 ± 0.02 & 0.23 ± 0.01 & 0.63 ± 0.01 & 0.61 ± 0.02 & 0.57 ± 0.07 \\
CheXpertPlus\_CheX\_MIMIC & 1.18 ± 0.04 & 0.20 ± 0.01 & 0.45 ± 0.01 & 0.59 ± 0.02 & 0.21 ± 0.01 & 0.62 ± 0.01 & 0.65 ± 0.02 & 0.58 ± 0.07 \\
RGRG & 1.17 ± 0.04 & 0.22 ± 0.01 & 0.44 ± 0.01 & 0.60 ± 0.02 & 0.22 ± 0.01 & 0.62 ± 0.01 & 0.67 ± 0.02 & 0.60 ± 0.07 \\
Cvt2distilgpt2\_MIMIC & 1.13 ± 0.04 & 0.20 ± 0.01 & 0.42 ± 0.01 & 0.61 ± 0.02 & 0.21 ± 0.01 & 0.61 ± 0.01 & 0.68 ± 0.02 & 0.61 ± 0.07 \\
RaDialog & 1.09 ± 0.03 & 0.20 ± 0.01 & 0.44 ± 0.01 & 0.54 ± 0.02 & 0.20 ± 0.01 & 0.59 ± 0.01 & 0.59 ± 0.02 & 0.54 ± 0.07 \\
CheXpertPlus\_MIMIC & 0.99 ± 0.04 & 0.18 ± 0.01 & 0.39 ± 0.01 & 0.59 ± 0.02 & 0.17 ± 0.01 & 0.58 ± 0.01 & 0.66 ± 0.02 & \textbf{0.62 ± 0.07} \\
BiomedGPT\_IU & 0.96 ± 0.05 & 0.14 ± 0.01 & 0.38 ± 0.01 & 0.52 ± 0.02 & 0.21 ± 0.01 & 0.54 ± 0.01 & 0.52 ± 0.02 & 0.54 ± 0.07 \\
CheXpertPlus\_CheX & 0.92 ± 0.03 & 0.16 ± 0.01 & 0.41 ± 0.01 & 0.49 ± 0.02 & 0.15 ± 0.01 & 0.53 ± 0.01 & 0.54 ± 0.02 & 0.55 ± 0.07 \\
VLCI\_MIMIC & 0.91 ± 0.04 & 0.14 ± 0.01 & 0.36 ± 0.01 & 0.48 ± 0.02 & 0.22 ± 0.01 & 0.58 ± 0.01 & 0.47 ± 0.02 & 0.49 ± 0.08 \\
CheXagent & 0.83 ± 0.05 & 0.12 ± 0.01 & 0.35 ± 0.01 & 0.49 ± 0.02 & 0.14 ± 0.01 & 0.50 ± 0.01 & 0.39 ± 0.03 & 0.57 ± 0.07 \\
GPT4V & 0.71 ± 0.03 & 0.08 ± 0.00 & 0.27 ± 0.01 & 0.41 ± 0.02 & 0.15 ± 0.01 & 0.52 ± 0.01 & 0.65 ± 0.03 & 0.55 ± 0.08 \\
LLM-CXR & 0.49 ± 0.02 & 0.03 ± 0.00 & 0.19 ± 0.01 & 0.06 ± 0.01 & 0.02 ± 0.00 & 0.28 ± 0.01 & 0.03 ± 0.01 & 0.30 ± 0.06 \\
\bottomrule
\end{tabular}
}

\end{table}

\begin{table}[!h]
\centering
\caption{Comprehensive evaluation of medical report generation models on the CheXpert Plus datasets. Models are ranked by 1/RadCliQ-v1.  Model evaluation results with 95\% confidence intervals (mean ± CI) under normality assumption. The best results for each metric are shown in \textbf{bold}.}
\label{tab:chexpert_leaderboard}
\vspace{6pt}

\renewcommand{\arraystretch}{1.2}
\setlength{\tabcolsep}{3pt}

\resizebox{\textwidth}{!}{
\begin{tabular}{lcccccccc}
\toprule
\rowcolor{gray!10}
\textbf{Model} & 
\textbf{1/RadCliQ-v1} $\uparrow$ &
\textbf{BLEU-2} $\uparrow$ & 
\textbf{BertScore} $\uparrow$ & 
\textbf{SembScore} $\uparrow$ & 
\textbf{RadGraph} $\uparrow$ & 
\textbf{RaTEScore} $\uparrow$ & 
\textbf{GREEN} $\uparrow$ & 
\textbf{1/FineRadScore} $\uparrow$ \\
\midrule
\multicolumn{9}{l}{\textbf{Findings + Impression}} \\
\midrule
CheXpertPlus\_CheX & \textbf{0.51 ± 0.07} & \textbf{0.14 ± 0.01} & \textbf{0.02 ± 0.02} & 0.38 ± 0.03 & 0.07 ± 0.02 & 0.49 ± 0.02 & 0.36 ± 0.05 & 0.35 ± 0.11 \\
CheXpertPlus\_CheX\_MIMIC & 0.51 ± 0.07 & 0.14 ± 0.01 & 0.01 ± 0.02 & \textbf{0.39 ± 0.03} & \textbf{0.07 ± 0.02} & \textbf{0.50 ± 0.02} & \textbf{0.38 ± 0.04} & \textbf{0.36 ± 0.12} \\
MedVersa & 0.49 ± 0.06 & 0.09 ± 0.01 & 0.01 ± 0.02 & 0.34 ± 0.03 & 0.05 ± 0.01 & 0.45 ± 0.02 & 0.33 ± 0.05 & 0.35 ± 0.11 \\
CheXpertPlus\_MIMIC & 0.48 ± 0.06 & 0.10 ± 0.01 & 0.00 ± 0.02 & 0.32 ± 0.03 & 0.05 ± 0.01 & 0.43 ± 0.02 & 0.29 ± 0.04 & 0.35 ± 0.11 \\
RadFM & 0.44 ± 0.05 & 0.07 ± 0.01 & -0.04 ± 0.02 & 0.23 ± 0.03 & 0.03 ± 0.01 & 0.39 ± 0.02 & 0.14 ± 0.03 & 0.34 ± 0.09 \\
GPT4V & 0.43 ± 0.05 & 0.06 ± 0.01 & -0.07 ± 0.02 & 0.21 ± 0.02 & 0.03 ± 0.01 & 0.39 ± 0.01 & 0.18 ± 0.04 & 0.33 ± 0.10 \\
\midrule
\multicolumn{9}{l}{\textbf{Findings}} \\
\midrule
CheXpertPlus\_CheX\_MIMIC & \textbf{0.81 ± 0.12} & 0.15 ± 0.03 & 0.34 ± 0.04 & \textbf{0.40 ± 0.05} & \textbf{0.21 ± 0.03} & \textbf{0.50 ± 0.03} & \textbf{0.27 ± 0.05} & 0.35 ± 0.18 \\
MAIRA-2 & 0.79 ± 0.10 & \textbf{0.16 ± 0.03} & \textbf{0.36 ± 0.03} & 0.35 ± 0.04 & 0.19 ± 0.03 & 0.48 ± 0.02 & 0.27 ± 0.05 & \textbf{0.35 ± 0.18} \\
CheXpertPlus\_CheX & 0.79 ± 0.10 & 0.15 ± 0.03 & 0.34 ± 0.03 & 0.38 ± 0.04 & 0.19 ± 0.03 & 0.49 ± 0.03 & 0.24 ± 0.05 & 0.34 ± 0.20 \\
MedVersa & 0.72 ± 0.10 & 0.13 ± 0.02 & 0.32 ± 0.03 & 0.34 ± 0.05 & 0.15 ± 0.02 & 0.47 ± 0.03 & 0.24 ± 0.05 & 0.34 ± 0.18 \\
RaDialog & 0.71 ± 0.09 & 0.13 ± 0.02 & 0.31 ± 0.03 & 0.35 ± 0.05 & 0.14 ± 0.02 & 0.45 ± 0.02 & 0.21 ± 0.04 & 0.33 ± 0.17 \\
RGRG & 0.67 ± 0.11 & 0.15 ± 0.02 & 0.32 ± 0.04 & 0.27 ± 0.05 & 0.14 ± 0.02 & 0.45 ± 0.03 & 0.22 ± 0.04 & 0.34 ± 0.17 \\
CheXpertPlus\_MIMIC & 0.66 ± 0.10 & 0.14 ± 0.02 & 0.29 ± 0.03 & 0.29 ± 0.05 & 0.13 ± 0.03 & 0.43 ± 0.03 & 0.24 ± 0.05 & 0.34 ± 0.19 \\
CheXagent & 0.64 ± 0.11 & 0.12 ± 0.02 & 0.28 ± 0.04 & 0.27 ± 0.04 & 0.12 ± 0.02 & 0.43 ± 0.02 & 0.18 ± 0.05 & 0.34 ± 0.19 \\
Cvt2distilgpt2\_MIMIC & 0.63 ± 0.08 & 0.12 ± 0.02 & 0.27 ± 0.03 & 0.27 ± 0.04 & 0.12 ± 0.02 & 0.42 ± 0.03 & 0.21 ± 0.05 & 0.35 ± 0.19 \\
VLCI\_MIMIC & 0.59 ± 0.09 & 0.12 ± 0.02 & 0.23 ± 0.03 & 0.25 ± 0.04 & 0.10 ± 0.02 & 0.38 ± 0.03 & 0.17 ± 0.05 & 0.33 ± 0.20 \\
Cvt2distilgpt2\_IU & 0.58 ± 0.11 & 0.08 ± 0.02 & 0.27 ± 0.03 & 0.15 ± 0.05 & 0.10 ± 0.02 & 0.38 ± 0.03 & 0.15 ± 0.05 & 0.33 ± 0.20 \\
RadFM & 0.57 ± 0.08 & 0.08 ± 0.02 & 0.23 ± 0.03 & 0.22 ± 0.04 & 0.08 ± 0.01 & 0.40 ± 0.03 & 0.10 ± 0.03 & 0.33 ± 0.16 \\
GPT4V & 0.57 ± 0.08 & 0.08 ± 0.01 & 0.21 ± 0.03 & 0.23 ± 0.04 & 0.08 ± 0.02 & 0.41 ± 0.02 & 0.15 ± 0.05 & 0.34 ± 0.20 \\
VLCI\_IU & 0.56 ± 0.09 & 0.11 ± 0.02 & 0.22 ± 0.03 & 0.17 ± 0.05 & 0.09 ± 0.02 & 0.42 ± 0.03 & 0.19 ± 0.05 & 0.34 ± 0.20 \\
BiomedGPT\_IU & 0.55 ± 0.08 & 0.02 ± 0.01 & 0.20 ± 0.03 & 0.24 ± 0.04 & 0.06 ± 0.01 & 0.35 ± 0.03 & 0.12 ± 0.04 & 0.32 ± 0.18 \\
LLM-CXR & 0.52 ± 0.06 & 0.04 ± 0.01 & 0.16 ± 0.02 & 0.21 ± 0.04 & 0.04 ± 0.01 & 0.32 ± 0.02 & 0.02 ± 0.01 & 0.29 ± 0.13 \\
\bottomrule
\end{tabular}
}

\end{table}

\section*{Disclosures}
O.H and J.M are founders and hold equity in Gradient Health, a private company focused on health data accessibility and availability for commercial research. Gradient Health provided the Private Dataset used in this work and did not provide funding for this research and had no role in its design, execution, or publication. The Private Dataset's 4x downsampled version is available under the \textbf{\href{https://gradienthealth.io/terms-of-use/}{Gradient Health Public License}}.

\clearpage
\bibliographystyle{sn-mathphys} 
\bibliography{sn-bibliography} 

\end{document}